# INDUCTION, OF AND BY PROBABILITY

Larry Rendell
Department of Computer Science,
University of Illinois at Urbana-Champaign,
1304 West Springfield Avenue, Urbana, Illinois 61801

## ABSTRACT

This paper examines some methods and ideas underlying the author's successful *probabilistic learning systems (PLS)*. These systems have proven uniquely effective and efficient for generalization learning *(induction)* in heuristic search. Aspects of PLS include use of probabilities to guide both task performance and learning, incremental revision and normalization of probabilities, and localization and correction of their errors. Construction of new terms (features) for heuristic functions may be feasible.

## 1. INTRODUCTION

Generalization learning or *induction* is the compression of data into some concise form allowing prediction [An 73, Mi 83, Re 85b, Wa 69]. This paper concerns the problem of *practical mechanized induction*, particularly as applied to heuristic search. *Efficiency* and *effectiveness* are the essence of practical induction.

**Inductive Difficulty.** For several reasons, induction is computationally difficult. First, the number of possible generalizations is very large, compared with the number of explicit hypotheses which can be explored in reasonable time. Secondly, available data are usually sparse, and so candidate hypotheses are hard to distinguish. Finally, data are often incorrect, or *noisy*, necessitating probabilistic treatment [Wa 69].

**Dual Role of Probability.** Probabilistic methods were used in early AI programs, some of which induced *heuristics* for *state-space search* [Ne 65, Sa 63, Sa 67]. These programs utilize probability or a related measure in two ways. First, the probability p that an object *(state)* contributes to success in the task domain becomes the basis for heuristic search there — the evaluation function H estimates p. Secondly, information about the nature of the variation of p with features of a state can help determine the parameters and form of H — i.e. p also plays a part in "higher level" inductive search. These roles of p alternate in an iterative, two stage algorithm: interpolated p values constrain task search, then observed p values constrain inductive search.

**PLS Capability.** This paper describes some advances implemented in the author's *probabilistic learning systems (PLS)* [Re 83a, Re 83c, Re 85c, Re 85d]. These systems are capable of efficient and effective incremental learning in noisy domains. The original program *PLS1* has achieved unique results, converging to optimal search heuristics when given features (attributes) by the user. A newer, radically different system *PLS0* constructs features for heuristic functions.

**Organization of this Paper.** The next section of this paper describes the problem. The third section outlines some PLS methods for incremental learning of probability, including normalization of biased data, and localization of errors. The fourth section considers probability as a factor in effective and efficient induction.

## 2. INDUCTIVE LEARNING

Induction is the formation of meaningful abstractions called *classes, categories,* or *concepts,* from data called *events, patterns,* or *objects.* An object might be a visual grid, the state of a checker board, or countless other items of interest, usually involving performance of some task. Objects within a class are *similar* for purposes of the current task. Because a class description embodies not only observed objects, but also similar objects yet to be encountered, induction is predictive.

**Deterministic Induction.** In a *feature space* representation,[1] an object is a vector $x = (x_1, x_2, ..., x_n)$, where n is the number of *features* $x_i$.[2] In vision for example, these features might be the light intensity (gray level) for a total of n squares (pixels) of a grid representing an image. If images contain some symbol of interest, each vector observed becomes a positive or negative instance of the concept "contains the symbol".

In a game such as checkers, the ultimate concept is "winning positions". Objects could be expressed as vectors of features such as piece advantage, center control, etc., or, instead, objects might be represented more basically as *states,* i.e. as vectors describing the contents of individual squares (in this case n = 32). Especially when the description vector is as detailed as this, observations cover just a fraction of feature space.

In *supervised* learning, observations provide positive and negative *training examples* of a concept or class. Fig. 1a shows a simple example, where the class is quite regular, and missing instances are predicted by an easy form of induction (insertion of *decision boundaries* or *clustering* in feature space [An 73, Du 73, To 74]).

**Probabilistic Induction.**[3] Instead of just two categories (positive and negative), the presence of uncertainty requires gradation, expressible using multiple classes as in Fig. 1b. These classes might represent a range of *probabilities* p. Associated with each p is some set $S_p$ of feature values; p can be considered the *name* of class $S_p$. $S_p$ often has some concise description, as in Fig. 2, where the leftmost cell is the *hyperrectangle* $r = S_p$, compactly expressible as $\{0 \leq x_1 \leq 4\} \cap \{0 \leq x_2 \leq 2\}$. Here $r$ represents the class having probability $p(r) = 0.2$ of occurrence of some event E, i.e. $p(r)$ is the conditional probability $Pr(E|r)$ of E, given that object $x \in r$.[4]

---

1. Objects and concepts must be expressed using some language, such as feature space representations, logic, grammars, etc. The full predicate logic cannot feasibly be used to induce complex relationships without restriction; on the other hand the limited feature vector representation cannot express structure without augmentation. In performing generalization using simple feature space representations, the best that can be done is to *cluster* neighboring points as shown in Fig. 1 [An 73]. See [Re 85a] for discussion and references about correspondences between languages.

2. In this paper we use the term *feature* to mean either a variable describing an object, or else the value of that variable. Similarly, the term *object* will refer either to the entity of interest, or else to some feature vector describing it.

3. Here we induce *probability classes*—i.e. classes of discrete probability values. We are NOT directly concerned with probability distributions of the classes as in [WB 68].

From one point of view, the number of classes is the number of p values; from another point of view, we have returned to the situation in which there are only two values (success or not). In the latter view, the feature space representation is a discrete function $p(r)$ indicating probability of inclusion in the "success" class.

**Structures and Methods for Probabilistic Induction (cells, centroids, and clustering).** Aside from associating success probabilities p with feature space cells $r$, there are auxiliary and alternative means of probabilistic induction (see [An 73, Du 73, To 74] for feature space methods and [Fu 82, Ka 72, Re 85d] for structured approaches). We could replace or accompany $r$ by other information. One possibility is a representative of $r$, its *prototype* or *centroid* $c$ [Du 73]. Another element is the error $e$ in $p$. Let us call an association of elements such as $(r, p, e, c)$ a *region* if it contains $r$ and $p$.

The author's learning systems are quite complex, e.g. their regions contain both cells and centroids. PLS1 uses cells for various reasons. Incremental learning demands progressively refined knowledge, and cells facilitate refinement (see Fig. 5, which will be discussed later). Rectangles $r$ are easy to express, and convey meaning well: conjunctions of feature ranges define classes (Fig. 2). Further, the size of $r$ in a region $R = (r, p, e, c)$ gives an impression of the magnitude of $\nabla p$ near $c$.

But prototypes may also be important. PLS uses centroids $c$ for several reasons. This vector $c$, along with the p value of a region R, define one datum in a regression to produce a heuristic for interpolated state evaluation (see below). In a sophisticated mode of system operation, $c$ allows measurement of the proximity of a state in feature space (the proximity measure weights features according to their "importance" near $c$ [Re 83d]).

PLS performs several operations on regions, including reclassification (of probability category p), differentiation (refinement or splitting of cells $r$), generalization (merging cells $r$), and reorganization (of $\{r\}$ [Re 85c]).

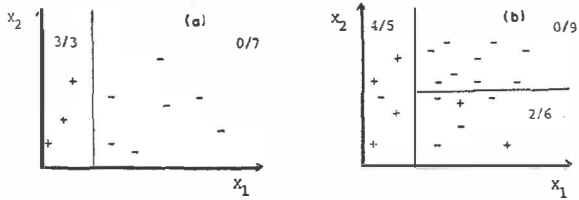

Figure 1. Deterministic versus probabilistic induction ($n = 2$). Ideally, classes can be neatly differentiated into positive and negative instances (a). But more commonly, exceptions occur (b). In less extreme cases, these are just anomalies whose effects can be recorded using proportions.

Because PLS is complex, we shall consider just part of its structure and just two of its operators: cell splitting and probability reclassification. Some other aspects will be mentioned briefly; the reader is referred to the bibliography for details. In the most of this paper we abbreviate regions to triples $(r, \hat{p}, e)$, where $\hat{p}$ represents a probability estimate within $r$, and e is the error in $\hat{p}$.

PLS1 splits cells using a clustering algorithm (described later in detail). Clustering criteria often measure object similarity only as a function of features, but this can cause problems [An 73]. Criteria based on something other than features are *external* criteria [An 73, p. 194]. Several years ago the author suggested *utility similarity* as a suitable external criterion when clustering relates to performance of some task [Re 76, Re 83a] (cf. [Go 78]). Utility similarity involves the whole *task environment*, not just features of an individual object. Here we measure utility as probability of success (Fig. 3).

Utility clustering is unlike information-theoretic clustering such as [WB 68], but similar in some respects to the methods of [Go 78, Qu 83]. We shall return to the topic of clustering later, to discuss the basic PLS1 method (§ 3), and to outline higher-dimensional clustering used by PLS0 for creation of structure (§ 4).

**Probabilistic Induction in Heuristic Search.** Let us consider a nondeterministic application. In heuristic search, noise arises because of inadequate features, search anomalies, and changing environments.

If rectangle $r$ contains a vector $x$ which represents a state in a problem or a game, "success" means leading to discovery of a good solution or win. Then the function $H(x) = p(r)$ is the probability that state $x \in r$ will be useful (its utility). H may be used as a discrete *evaluation function* or *heuristic* to assess state $x$ (see Figs. 2, 3).[5]

Instead of a discrete function, an alternative form for the heuristic is the linear combination $H(x) = b \cdot x$, where b is the *coefficient vector*. Regions $(r, \hat{p}, e, c)$ are used as data in weighted stepwise regression to determine $b$.[6] The author's probabilistic learning system PLS1 uses both of these forms [Re 83a, Re 83c, Re 85c], and also piecewise linear functions [Re 83d].

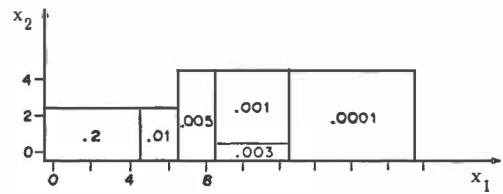

Figure 2. Probability classes. Feature space can be partitioned (clustered) into rectangular cells $r$, each having roughly similar probability p of occurrence of some event E (e.g. Fig. 3). The p values shown inside each $r$ represent $Pr(E|r)$. The set $\{(r, p)\}$ can be considered a step function of feature vector $x$ whose values are p. If used for heuristic search, this set of probability classes forms a discrete evaluation function.

## 3. PROBABILITY AS PRODUCT OF LEARNING

The probabilistic learning system PLS can be used for "single concept" learning, like the systems described in [Di82], but most work to date has been in heuristic search, which is hard because of inherent noise and various complications of incremental learning. Results using PLS1 include efficient convergence to optimal evaluation functions H. Novel aspects of PLS include incremental revision (reclassification) of probability estimates $\hat{p}(r)$ during task performance, appropriate refinement of feature space cells r, and normalization of heavily biased samples. These and other aspects are examined below.

**Definition and Use of a posteriori Probability.** In the supervised learning of a single class, the definition of success probability is clear: a training sample is in the class or not, and therefore contributes **1** or **0** in a success count for its corresponding feature space cell r. In problem solving, however, there are complications.

The PLS method for learning H begins as follows. Given an (initially trivial) H and a set S of sample problems, attempt their solution, to compute a corresponding set T of search trees. (There must be at least one solution in T if any useful information is to be extracted.) For feature space cell r, the probability of success p(r) is the number of states within r found on solution paths in T, divided by the total number of states within r.[7] These conditional probabilities $p(r) = Pr(success|r)$ are illustrated in Fig. 3 (see also Figs. 1, 2).

An ideal measure of state quality would be p(r) for *undirected (breadth-first)* search, *very large, random* S, and *very small* r. However, resource constraints preclude all three ideals. One alternative is to obtain probabilities using informed and progressively improving heuristics H, then to normalize to standard, breadth-first values. This involves probability estimates $\hat{p}(r)$, initially obtained for easy S and larger r, then for progressively harder S's and gradually refined r's. By making use of improving $\hat{p}(r)$ values, H may be bootstrapped. This whole approach, however, has several difficulties.

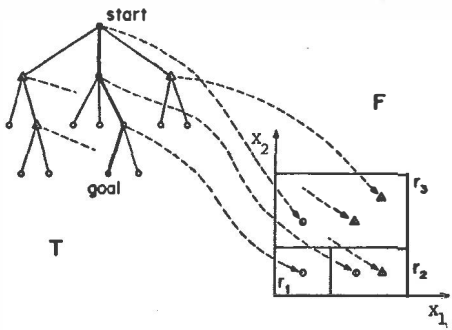

Figure 3. Probabilities computed from a search tree T. Nodes developed in T are mapped into feature space F. Here the arbitrary rectangular partition gives three values: $p(r_1) = 1/1 = 1.0$, $p(r_2) = 1/2 = 0.5$, and $p(r_3) = 1/3 = 0.3$. These pairs or *regions* $(r_i, p_i)$ may be used to define a discrete heuristic function, or they may become data to fit a regression function, to represent the probability of a node's being on a solution path.

---

7. A similar probability is defined for games in [Re83c]. See also [Pe83].

---

**Good Use of Scarce Information To Estimate Probabilities.** One purpose of induction is to minimize knowledge acquisition costs, so training samples x are usually sparse. These data alone are not sufficient, so a *model* is used. Models may take various forms; two suitable for heuristic search have already been mentioned. One is the linear combination $H(x) = Pr(success|x) = \mathbf{b.x}$. A more general model is the discrete function $H(r) = \hat{p}(r)$ of feature space cells r.

For either model to be useful, $\hat{p}$ (or any other basis for H) must vary smoothly with x. Then sizable volumes of feature space are meaningful as units, and neighboring coordinates x may be *clustered* into cells r.[8] PLS clusters similar utilities (here success probabilities — see §2).[9]

The system employs a splitting algorithm which repeatedly dichotomizes rectangles r using a *dissimilarity* measure d. If $p_1$ and $p_2$ are the two probabilities for a tentative dichotomy, and $e_1$ and $e_2$ their error *factors*, then $d = |\log p_1 - \log p_2| - \log(e_1 e_2)$. This measure is computed for a number of boundary insertions parallel to feature space axes. If the largest d is positive, the corresponding split is retained. The process is repeated until additional refinement is unwarranted by the data (until $d \le 0$). Larger d means more *assured* dissimilarity.

First used in PLS1 several years ago [Re76], the task-utility- or probability-based dissimilarity d is somewhat like the information-theoretic measure d' of [Qu83].[10] To relate the two measures intuitively: splitting governed by utility difference d lessens our ignorance (of the probability of success) [Wa69]. Our clustering criterion d is unlike information-theoretic measures such as [WB68], which consider probability distributions of classes. We do not need to know the distributions of probability values p, although their errors e relate.

---

8. Whenever the task utility (here measured as $\hat{p}$) varies smoothly with features, induction is relatively simple, or *selective* (cf. [Mi83]). Many algorithms rely on smoothness for efficiency. Examples include discriminant functions and cluster analysis [An73, Du73, To74], and systems using the restriction of predicate logic $VL_1$ [Mi83], which corresponds to our rectangular feature space cells. The overriding concern in such algorithms is gradual change of utility (probability) with feature values. For example, in checkers, the quality of a state (probability of its contributing to winning) varies smoothly and monotonically with the abstract features typically used, such as piece advantage, mobility, center control, etc.
Ideally, a learning system would be able to deal with more primitive features which describe objects in detail. In checkers, primitives would be elementary board descriptions giving contents of the 32 squares which can be legally occupied. However, no mechanized learning system has this capability. In an analysis of the difficulty involved, [Re83b, Re83c] discuss this example as a case requiring *transformation of variables* [To74], also known as the *problem of new terms* [Di82], or *constructive induction* [Mi83]. New descriptions (concepts) must be created which amount to the usual high level features [Re85d] examines a layered method for reducing the complexity of this problem, and describes some encouraging experiments. We shall return to this topic in §4.

9. Besides clustering, which here is a splitting or specialization rule, PLS uses other rules which generalize or rearrange feature space cells [Re85c].

10. $d' = q_1 \log q_1 + q_2 \log q_2 + m_1 \log m_1 + m_2 \log m_2$, where, in terms of our p's above, $q_i = p_i/(p_1 + p_2)$, and $m_i = (1-p_i)/(2 - p_1 - p_2)$.

Clustering similar probabilities improves accuracy since it effectively produces larger samples. Since data determine splits, the sizes and shapes of feature space cells (clusters) tend to match characteristics of the domain (see Fig. 2).

PLS may be compared with Samuel's statistical method of adjusting parameters, and with his signature tables [Sa 63, Sa 67]. All these schemes use available information well, allowing every state encountered to influence H stochastically — each useful state increments a "success" count (see Fig. 3). Consequently, PLS and Samuel's checker player are both effective and efficient. PLS is more stable, general, and mechanized [Re 85c].

**Incremental Learning of Probability Estimates.** Initially, PLS may possess no heuristic information. To start, easy problems can be presented which are solvable within resource constraints (Fig 4).[10]

The data so obtained are assumed to represent all problems, and the resulting probabilities are used predictively, to form heuristic $H(x) = b \cdot x$ or $H(r) = \hat{p}(r)$ for a new round of problem solving or game playing. Because H is now improved, harder problems can be solved, and consequently, new probability measurements $p(x)$ become available (called *elementary* probabilities). These are used both to improve (reclassify) $\hat{p}(r)$, and also to refine (split) current cells r (Fig. 5). These two operations are detailed below, but first we shall deal with a phenomenon arising from the incremental nature of PLS.

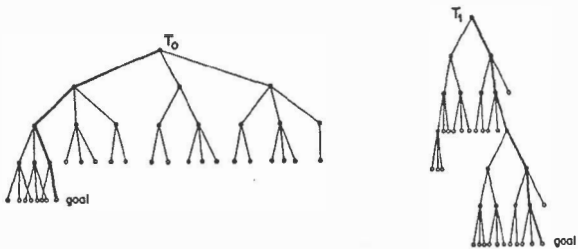

Figure 4. Effects of improving heuristic functions. As knowledge increases over successive iterations of the system, search becomes narrower. $T_1$ develops as many nodes as $T_0$ (each 15), but has 50% more "success" nodes. In terms of information available for learning, this means that "success" counts are higher (a greater proportion of the nodes appear on the solution path or game win), which aids learning. However, the same phenomenon biases the probabilities unpredictably.

**Normalization of Inherently Biased Probabilities.** When the system computes probabilities $p(x)$, it generally uses a non-trivial heuristic H to sample solutions to difficult problems. These $p(x)$ are heavily biased, they overestimate true success probability. Positive bias occurs because fewer useless states are developed when H guides search, and a higher proportion participate in solutions (see Fig 4).

The elementary values $p(x)$ must be corrected if commensurate state evaluation and meaningful probability learning are to be possible. Unfortunately, while this bias is known to be positive, its magnitude is unpredictable. Nevertheless, elementary probabilities may be normalized, coarsely in any one system iteration, but effectively overall. Incremental feedback over repeated iterations of the learning system serves to correct errors in rough treatments. One method (a simplification of [Re 83a]) is depicted in Fig. 6. Methods of error detection and correction will be discussed shortly.

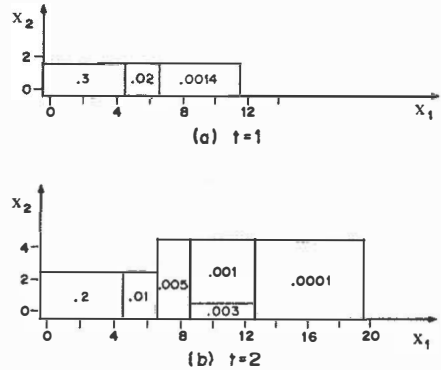

Figure 5. Incremental growth of feature space cells. As information accumulates with continuing task performance, the partition becomes more refined. The volume of feature space covered gradually extends to enclose all data encountered over repeated iterations of the system.

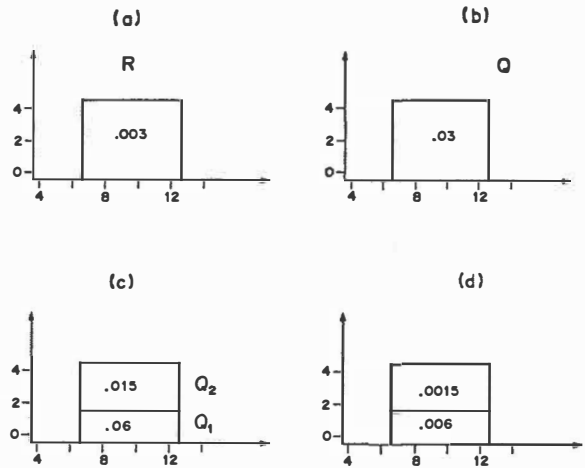

Figure 6. One simplified method of normalizing probability estimates. The *cumulative* region R in (a) estimates "true" (unbiased) success probability. The *elementary* region Q in (b) obtained using informed search has a rectangle matching that of R. The ratio of the two probabilities $\hat{p}_R / p_Q = .03/.003 = 10$. A hypothetical outcome of splitting Q is diagrammed in (c). To normalize $Q_1$ and $Q_2$, the elementary probabilities can simply be divided by 10, to give (d). This is a simplification of one of the normalization procedures of [Re 83a].

---

**Refinement of Feature Space Cells with Improving Probability Resolution.** As more training problems are encountered in successive PLS iterations, domain knowledge accumulates in a *region set*, a set of triples $(r, \hat{p}, e)$ (recall from §2 that e is the error in $\hat{p}$). Region set refinement uses the clustering algorithm previously described. Each *established* or *cumulative* region becomes the starting point for further splitting (Fig. 5).

This learning is not just linear accretion of information, but rather an accelerating process guided by experience. Cells r have size and shape determined by the particular domain. These cells become appropriately refined (split by the clustering algorithm) whenever significant differences emerge in success probability. Further, as the heuristic H improves, it guides search toward more successful states, and this results in more useful information. More "success" states permit greater contrast in probability (see Figs. 4, 5, 6). The iterative process can be compared to search by a telescope: at first a large area might be scanned, and when something interesting appears, magnification is increased, now omitting surrounding incidentals, and allowing greater resolution of the phenomenon of interest. In our case, the general phenomenon is variation of $\hat{p}$ with its determinants, the features, and the phenomenon of particular interest is variation of $\hat{p}$ in volumes of feature space having fast-changing $\hat{p}$ values.

**Improvement of Probability Estimates and Localization of Inaccuracies (discovery and correction of errors).** In PLS1, the region set $\{(r, \hat{p}, e)\}$ is the fundamental knowledge structure, and information accumulates in this set. PLS has several means of detecting errors e and improving probability estimates $\hat{p}$. The simplest method of dealing with errors is to moderate their influence by statistical regression — fitting a linear model $H(x) = b \cdot x$, using the regions $\{(r, \hat{p}, e)\}$ as data to find b. This is repeated at every iteration, with the region set still the only reservoir of accumulating heuristic knowledge. Regression has the effect of smoothing errors in probabilities $\hat{p}$, which is very important, since one high estimate can vitiate a discrete heuristic function (a secondary effect of this regression is the increased heuristic power resulting from the interpolation $H(x) = b \cdot x$).

Another means of improving probability estimates involves the incremental creation of fresh data $\{(x, p', e')\}$ in every new iteration. These data are used to generate a (normalized) region set $\{(r, \hat{p}', e')\}$ corresponding to the already available, established set $\{(r, \hat{p}, e)\}$ (see Fig. 6). For each established rectangle r, the two probability estimates $\hat{p}$ and $p'$ are averaged and the error estimate resulting from combining e and e' is decreased.[11]

Finally, the most sophisticated kind of error correction involves the genetic learning system *PLS2* which controls parallel activations of the basic system PLS1. PLS2 is able to locate errors in individual regions. This scheme is stable, efficient and effective [Re 83c, Re 85c].

In summary, PLS1 and PLS2 use clustering iteratively, to refine knowledge gradually. Because of the progressive improvement in the resulting heuristic, all iterations but the first generate biased probabilities, which must be normalized. The errors involved are large and difficult to estimate, but can be improved through iterative feedback. The system is effective and efficient.

## 4. PROBABILITY AS INDUCTIVE CRITERION

PLS not only *learns* probability, it is also *guided by* probability. Consider a simple kind of induction, that of learning a single class C, given a set S of objects which are examples of C. This can be an immense task. If the potential number of discrete positive and negative examples of C is only 100 (e.g. in a two dimensional $10 \times 10$ feature space), the number of inductions (ways of forming C) is $2^{100} \simeq 10^{10}$.

As Watanabe showed in his "theorem of the ugly duckling", no one classification is intrinsically better than any other [Wa 69]. In other words, in order to select an appropriate class, we must rely on some *external* criterion. This criterion is the quality or *credibility;* it expresses some ascribed elegance or purpose of a generalization. The credibility may be considered a probability (that we trust or value class C).

**Two Kinds of Credibility.** Watanabe discusses and formalizes two kinds of credibility criteria: *evidential*, and *extra-evidential*. These are also called *confirmation* and *creditation*. Evidential and extra-evidential criteria may be combined, to form a single credibility measure.[12]

The simplest evidential criterion in our case involves the *performance* of an induced heuristic H, viz. the number of nodes developed in search for problem solutions. In PLS2, this does not just confirm H, it assigns credit and blame to individual regions [Re 83c, Re 85c]).

A related aspect of PLS1 involves the task-oriented success probability $\hat{p}$. Since we induce for the purpose of successful heuristic search, we want the product of induction (evaluation function H) to differentiate $\hat{p}$. Differences and similarities in $\hat{p}$ are the basis for the clustering algorithm described earlier.

One extra-evidential criterion used in PLS1 is a form of *simplicity:* feature space cells are rectangular.

**Credibility Criteria and the Complexity of Induction.** Both kinds of credibility can aid the extreme computational problem of induction. For example, the constraint that feature space cells be rectangular serves not only to *select* generalizations, but also to *speed* the inductive process. Smoothness in probability-feature relationships permits sizable cells to capture similarities in $\hat{p}$, and the use of rectangles reduces time complexity of algorithms.

Programs for practical induction improve efficiency by simplifying in various ways. Most begin with abstract features such as piece advantage, center control, etc., which affect the success probability $\hat{p}$ smoothly.

Induction is the realization of regularity, partly discovered and partly imposed. Different kinds of credibility criteria and other constraints may be implemented in various forms, all of which provide regularity and improve efficiency: some components become information structures, some are built into algorithms, and some are parameters of algorithms [Re 85a].

---

11. This process of averaging probability estimates uses error-weighting. See [Re 83a, Re 83c].

12. Evidential criteria correspond to the AI notion of data-drivenness, and extra-evidential criteria correspond to model-drivenness.

**Probabilistic Structural Induction.** As explained in Footnote 8, we have so far been dealing solely with selective induction, whereas the great challenge is constructive induction requiring substantial transformation of the original features (primitives).

The new system *PLS0* is designed to begin with very detailed primitives such as the contents of individual squares of a checker board. The variation of probability p with primitives is highly irregular, and methods like PLS1 are useless since they rely on uniformity of p within clusters. However PLS0 employs a novel *higher-dimensional* clustering, which groups not just features, or even simple probabilities, but rather probability *surfaces* in subspaces of primitives. These surfaces represent interrelationships among components of objects. Higher-dimensional clustering *creates structure* [Re 83c, Re 85d]. The system effectively induces rules of a grammar.

Extending the methods discussed in this paper, PLS0 imposes various constraints or credibility measures, and breaks down the problem into several levels, each with a reduced time complexity (e.g. polynomial instead of double exponential) [Re 85d].

## SUMMARY

This paper has outlined some methods used by the author's probabilistic learning systems for induction of heuristic functions. PLS systems cluster task utility in feature space. Utility may be a probability, used both to guide heuristic search and also to induce the heuristic itself. Among the PLS operators are normalization of necessarily biased probabilities, and discovery and correction of errors. In PLS0, a higher-dimensional extension of clustering may become part of a feasible method for inducing structure.

Efficiencies result from exploiting various constraints. The key to practical, efficient and effective induction may be found in certain aspects of the probabilistic variation of utility with features, particularly similarity, accuracy, and credibility. These may be incorporated in effective learning schemes which employ reclassification, refinement, reorganization, and layering. Many seemingly diverse learning systems have incorporated ideas discussed in this paper.

## ACKNOWLEDGEMENTS

I would like to thank Chris Matheus, Raj Seshu, and the Workshop reviewers for their helpful suggestions.